\documentclass[journal,10pt,twocolumn,twoside]{IEEEtran}
\bstctlcite{IEEEexample:BSTcontrol}
\usepackage{times,amsmath,amssymb}
\usepackage{graphicx}
\usepackage{setspace}
\usepackage[ruled]{algorithm2e}
\usepackage{algorithmicx}
\usepackage{algpseudocode}
\usepackage{pdfpages}

\title{Class Mean Vector Component and Discriminant Analysis}
\author{\IEEEauthorblockN{Alexandros Iosifidis}\\
\IEEEauthorblockA{Department of Engineering, ECE, Aarhus University, Denmark\\
Email: alexandros.iosifidis@eng.au.dk}
}
\normalsize
\begin{document}
\maketitle

\newpage

\begin{abstract}
The kernel matrix used in kernel methods encodes all the information required for solving complex nonlinear problems defined on data representations in the input space using simple, but implicitly defined, solutions. Spectral analysis on the kernel matrix defines an explicit nonlinear mapping of the input data representations to a subspace of the kernel space, which can be used for directly applying linear methods. However, the selection of the kernel subspace is crucial for the performance of the proceeding processing steps. We propose a new optimization criterion, leading to a new component analysis method for kernel-based dimensionality reduction that optimally preserves the pair-wise distances of the class means in the feature space. This leads to efficient kernel subspace learning, which is crucial for kernel-based machine learning solutions. We provide extensive analysis on the connections and differences between the proposed criterion and the criteria used in kernel Principal Component Analysis, kernel Entropy Component Analysis and Kernel Discriminant Analysis, leading to a discriminant analysis version of the proposed method. Our theoretical analysis also provides more insights on the properties of the feature spaces obtained by applying these methods. Results on a variety of visual classification problems illustrate the properties of the proposed methods.
\end{abstract}

\section{Introduction}\label{S:Intro}
Kernel methods are very effective in numerous machine learning problems, including nonlinear regression, classification, and retrieval. The main idea in kernel-based learning is to nonlinearly map the original data representations to a feature space of (usually) increased dimensionality and solve an equivalent (but simpler) problem using a simple (linear) method for the transformed data. That is, all the variability and richness required for solving a complex problem defined on the original data representations is encoded by the adopted nonlinear mapping. Since for commonly used nonlinear mappings in kernel methods the dimensionality of the feature space is arbitrary (virtually infinite), the data representations in the feature space are implicitly obtained by expressing their pair-wise products stored in the so-called kernel matrix $\mathbf{K} \in \mathbb{R}^{N \times N}$, where $N$ is the number of samples forming the problem at hand.

The feature space determined by spectral decomposition of $\mathbf{K}$ has been shown to encode several properties of interest: it has been used to define low-dimensional features suitable for linear class discrimination \cite{yang2005kPCA}, to train linear classifiers capturing nonlinear patterns of the input data \cite{muller2001introduction,iosifidis2016multiclass}, to reveal nonlinear data structures in spectral clustering and diffusion maps \cite{shi2000normalized,Coifman7426} and it has been shown to encode information related to the entropy of the input data distribution \cite{jenseen2010keca}. The expressive power of $\mathbf{K}$ and its resulting basis has also been used in problems requiring discriminative learning \cite{plawiak2019novel,abdar2019new,wang2017sparse,dong2017person,esmaeilzehi2017nonparametri,das2017exploring,iosifidis2015kernel,plawiak2020dghnl,plawiak2019application}, regression \cite{jebaduari2017sksvr}, representation learning \cite{maggu2017kernel} and transfer learning \cite{xue2017transfer}. However, the selection of the kernel subspace is crucial for the performance of the proceeding processing steps, as discarding information related to the problem at hand at the initial processing steps would lead to low performance.

In this paper we first propose a kernel matrix component analysis method for kernel-based dimensionality reduction optimally preserving the pair-wise distances of the class means in the kernel space. We show that proposed criterion also preserves the distances of the class means with respect to the total mean of the data in the kernel space, as well as the Euclidean divergence between the class distributions in the input space. We analyze the connection of the proposed criterion with those used in (uncentered) kernel Principal Component Analysis (kPCA), kernel Entropy Component Analysis (kECA) and Kernel Discriminant Analysis (KDA), providing new insights related to the dimensionality selection process of these two methods. KPCA and kECA select the eigen-pairs corresponding to the maximal eigenvalues or entropy values, respectively. As we will show in the following, for the selection of an eigen-pair in the proposed method, called Class Mean Vector Component Analysis (CMVCA), both the eigenvalue needs to be high and the corresponding eigenvector needs to be angularly aligned to the difference of a pair of class indicator vectors. Finally, exploiting the connection of the proposed method to KDA, we propose a discriminant analysis method, called Class Mean Vector Discriminant Analysis (CMVDA), that is able to produce kernel subspaces the dimensionality of which is not bounded by the number of classes forming the problem at hand. Experiments on real-world data illustrate our findings.

The contributions of the paper are the following:
\begin{itemize}
    \item We propose a new criterion of defining the subspace of the kernel space.
    \item We provide extensive theoretical analysis of the proposed criterion highlighting its properties in terms of preserving the scatter of the data from the dataset mean, the Euclidean divergence between the class probability density functions, and the class discrimination when measured by the Rayleigh quotient criterion.
    \item We provide an extensive analysis of the connections and differences between the proposed criterion and those used in (uncentered) kPCA, kECA and KDA, leading to the proposed CMVDA.
    \item We show how the proposed criterion can be efficiently combined with kernel approximation and randomization methods for performing kernel subspace learning on large data sets. 
\end{itemize}

The remainder of the paper is structured as follows. Section \ref{S:Preliminaries} provides the theoretical foundation needed for the analysis in our work. Section \ref{S:ProposedMethod} describes in detail the proposed Class Mean Vector Component Analysis. It provides analysis of its properties, and its connections and differences with the related techniques. Section \ref{SS:DiscriminantCVMA} describes the Class Mean Vector Discriminant Analysis, an extension of the proposed criterion to discriminant kernel subspace learning. Section \ref{SS:CMVAapproxKernels} describes how the proposed methods can be efficiently combined with kernel approximation and randomization methods for performing kernel subspace learning on large data sets. An experimental study illustrating the properties of the proposed methods, in comparison with the related methods, is provided in Section \ref{S:Experiments}. Finally, conclusions are drawn in Section \ref{S:Conclusions}.

\section{Preliminaries}\label{S:Preliminaries}
Let us denote by $\mathcal{S} = \{\mathcal{S}_c\}_{c=1}^{C}$ a set of $D$-dimensional vectors, where $\mathcal{S}_c = \{\mathbf{x}^{c}_{1},\dots,\mathbf{x}^{c}_{N_c}\}$ is the set of vectors belonging to class $c$. In kernel-based learning \cite{muller2001introduction}, the samples in $\mathcal{S}$ are mapped to the kernel space $\mathcal{F}$ by using a nonlinear function $\phi(\cdot)$, such that $\mathbf{x}_i \in \mathbb{R}^{D} \rightarrow \phi(\mathbf{x}_i) \in \mathcal{F}, \:i=1,\dots,N$, where $N = \sum_{c=1}^C N_c$. Since the dimensionality of $\mathcal{F}$ is arbitrary (virtually infinite), the data representations in $\mathcal{F}$ are not calculated. Instead, the non-linear mapping is implicitly performed using the kernel function expressing dot products between the data representations in $\mathcal{F}$, i.e. $\kappa(\mathbf{x}_i, \mathbf{x}_j) = \phi(\mathbf{x}_i)^T \phi(\mathbf{x}_j)$. By applying the kernel function on all training data pairs, the so-called kernel matrix $\mathbf{K} \in \mathbb{R}^{N \times N}$ is calculated. One of the most important properties of the kernel function $\kappa(\cdot,\cdot)$ is that it leads to a positive semi-definite (PSD) kernel matrix $\mathbf{K}$. While the use of indefinite matrices \cite{scheif2015indefinite,gisbrecht2015metric} and general similarity matrices \cite{balcan2008theory} have also been researched, in this paper we will consider only positive semi-definite kernel functions.

The importance of kernel methods in Machine Learning comes from the fact that, in the case when a linear method can be expressed based on dot products of the input data, they can be readily used to devise nonlinear extensions. This is due to the Representer theorem \cite{muller2001introduction} stating that the solution of a linear model in $\mathcal{F}$, e.g. $\mathbf{W}_{(\phi)} \in \mathbb{R}^{|\mathcal{F}| \times M}, \: M \le min(D,N)$, can be expressed as a linear combination of the training data, i.e. $\mathbf{W}_{(\phi)} = \mathbf{\Phi} \mathbf{A}$, where $\mathbf{\Phi} = [\phi(\mathbf{x}_1),\dots,\phi(\mathbf{x}_N)]$ and $\mathbf{A} \in \mathbb{R}^{N \times M}$ is a matrix containing the combination weights. Then, the output of a linear model in $\mathcal{F}$ can be calculated by $\mathbf{o}_i = \mathbf{W}_{(\phi)}^T \phi(\mathbf{x}_i) = \mathbf{A}^T \mathbf{k}_i$, where $\mathbf{k}_i \in \mathbb{R}^N$ is a vector having its $j$-th element equal to $[\mathbf{k}_i]_j = \kappa(\mathbf{x}_j,\mathbf{x}_i), \:j=1,\dots,N$. That is, instead of optimizing with respect to the arbitrary dimensional $\mathbf{W}_{(\phi)}$, the solution involves the optimization of the combination weights $\mathbf{A}$. Another important aspect of using kernel methods is that they allow us to train models of increased discrimination power \cite{muller2001introduction,vapnik1998statistical}. Considering the Vapnik-Chervonenkis (VC) dimension of a linear classifier defined on the data representations in the original feature space $\mathbb{R}^D$, the number of samples that can be shattered (i.e., correctly classified irrespectively of their arrangement) is equal to $D+1$. On the other hand, the VC dimension of a linear classifier defined on the data representations in $\mathcal{F}$ is higher. For the most widely used kernel function, i.e. the Gaussian kernel function $\kappa(\mathbf{x}_i,\mathbf{x}_j) = exp\left( -\frac{1}{2\sigma^2} \|\mathbf{x}_i - \mathbf{x}_j\|_2^2\right)$, it is virtually infinite. In practice this means that, under mild assumptions, a linear classifier applied on data representations in $\mathcal{F}$ can classify all training data.

Using the definition of the kernel matrix, i.e. $\mathbf{K} = \mathbf{\Phi}^T \mathbf{\Phi}$, and its PSD property, we can determine a subspace of the corresponding kernel space $\mathcal{F}$. This can be done by its spectral decomposition $\mathbf{K} = \mathbf{U} \mathbf{\Lambda} \mathbf{U}^T$, leading to $\tilde{\mathbf{\Phi}} = [\mathbf{\phi}_1,\dots,\mathbf{\phi}_N] = \mathbf{\Lambda}^{\frac{1}{2}} \mathbf{U}^T$, where $\mathbf{\Lambda} = \textrm{diag}(\lambda_1,\dots,\lambda_N)$ and $\mathbf{U} \in \mathbb{R}^{N \times N}$ are the eigenvalues and the corresponding eigenvectors of $\mathbf{K}$. Thus, an explicit nonlinear mapping from $\mathbf{x}_i \in \mathbb{R}^D$ to $\mathbf{\phi}_i \in \mathbb{R}^{N}$ is defined, such that the $d$-th dimension of the training data is:
\begin{equation}
[\tilde{\mathbf{\Phi}}]_d = \sqrt{\lambda_d} \mathbf{u}_d^T, \label{Eq:Phi_d}
\end{equation}
where $\lambda_d$ is the $d$-th largest eigenvalue of $\mathbf{K}$ and $\mathbf{u}_d$ is the corresponding eigenvector. In the case where $\mathbf{K}$ is centered in $\mathcal{F}$, $\mathbb{R}^{N}$ is the space defined by kernel PCA \cite{muller2001introduction}. Moreover, as has been shown in \cite{kwak2013npt,kwak2017IncrNPT}, the kernel matrix needs not to be centered. $N$ is called the effective dimensionality of $\mathcal{F}$ and $\mathbb{R}^N$ is the corresponding effective subspace of $\mathcal{F}$. As can be observed by Eq. (\ref{Eq:Phi_d}), the effective dimensionality of $\mathcal{F}$ depends on the rank of $\mathbf{K}$, as for eigenvectors corresponding to zero eigenvalues, the corresponding subspace dimensions collapse to zero. This is essentially the same as the uncentered kernel PCA. The kECA was proposed \cite{jenseen2010keca} following the uncentered kernel approach and sorting eigenvectors based to the size of the entropy values defined as $(\sqrt{\lambda_d}\mathbf{u}_d^T\mathbf{1})^2$. KECA has also been shown to be the projection that optimally preserves the length of the data mean vector in $\mathcal{F}$ \cite{jenseen2013mean}.

After sorting the eigen-vectors based on the size of either the eigenvalues, or the entropy values, the $l$-th dimension of a sample $\mathbf{x}_j$ in the kernel subspace is obtained by:
\begin{equation}
[\mathbf{y}_j]_l = \lambda_l^{-\frac{1}{2}} \mathbf{u}_l^T \mathbf{k}_j,
\end{equation}
where $\mathbf{k}_j \in \mathbb{R}^N$ is a vector having elements $[\mathbf{k}_j]_i = \kappa(\mathbf{x}_i,\mathbf{x}_j), \:i=1,\dots,N$. Note that the use of such an explicit mapping preserves the discriminative power of the kernel space, since a linear classifier on the data representations in $\mathbb{R}^N$ can successfully classify all $N$ training samples.

When a lower-dimensional subspace $\mathbb{R}^M, \:M < N$ of $\mathcal{F}$ is sought, the criterion for selecting an eigen-pair ($\mathbf{u}_d,\lambda_d$) is defined in a generative manner, i.e. minimizing the quantity $\|\mathbf{K} - \mathbf{U}_M \mathbf{\Lambda}_M \mathbf{U}_M^T\|_2^2$ leading to selecting the eigen-pairs corresponding to the $M$ maximal eigen-values $\lambda_1 \ge \lambda_2 \ge \dots \ge \lambda_M$ in the case of kPCA, or maximizing the entropy of the data distribution leading to selecting the eigen-pairs corresponding to the $M$ maximal entropy values in the case of kECA.

\section{Class Mean Vector Component Analysis}\label{S:ProposedMethod}
Since the data representations in the kernel space $\mathcal{F}$ form classes which are linearly separable, we make the assumption that classes in $\mathcal{F}$ are unimodal. We express the distance between classes $k$ and $m$ by:
\begin{equation}
d(c_k,c_m) = \|\mathbf{m}_k - \mathbf{m}_m\|_2^2, \label{Eq:TwoVectorsDistancePhi}
\end{equation}
where $\mathbf{m}_c$ is the mean vector of class $c_c$ in $\mathcal{F}$. Since $d(c_k,c_m)$ is calculated by using elements of the kernel matrix $\mathbf{K}$, i.e. $d(c_k,c_m) = [\mathbf{k}_k]_k - 2 [\mathbf{k}_k]_m + [\mathbf{k}_m]_m$, we exploit the spectral decomposition of $\mathbf{K}$ and express the mean vectors in the effective kernel subspace, i.e., $\mathbf{m}_c \in \mathbb{R}^N$ with their $d$-th dimension equal to:
\begin{equation}
[\mathbf{m}_c]_d = \frac{1}{N_c} \sum_{\mathbf{\phi}_i \in \mathcal{S}_c} [\mathbf{\phi}_i]_d = \sqrt{\lambda_d} \mathbf{u}_d^T \mathbf{e}_c,
\end{equation}
where $\mathbf{e}_c \in \mathbb{R}^N$ is the indicator vector for class $c$ having elements equal to $[\mathbf{e}_c]_i = 1 / N_c$ for $\mathbf{\phi}_i \in \mathcal{S}_c$, and $[\mathbf{e}_c]_i = 0$ otherwise. Then, $d(c_k,c_m)$ takes the form:
\begin{eqnarray}
d(c_k,c_m) &=& \sum_{d=1}^N \lambda_d \left( \mathbf{u}_d^T \mathbf{e}_k - \mathbf{u}_d^T \mathbf{e}_m \right)^2 \nonumber \\
&=& \sum_{d=1}^N \lambda_d \left( \mathbf{u}_d^T (\mathbf{e}_k - \mathbf{e}_m) \right)^2 \nonumber \\
&=& \sum_{d=1}^N \lambda_d \left( \|\mathbf{u}_d\|_2 \: \|\mathbf{e}_k - \mathbf{e}_m\|_2 \: cos( \mathbf{u}_d, \mathbf{e}_k - \mathbf{e}_m ) \right)^2 \nonumber \\
&=& \frac{N_k + N_m}{N_k N_m} \sum_{d=1}^N \lambda_d \cos^2( \mathbf{u}_d, \mathbf{e}_k - \mathbf{e}_m ). \label{Eq:MeanVectorsDistancePairs}
\end{eqnarray}
From the above, it can be seen that the eigen-pairs of $\mathbf{K}$ maximally contributing to the distance between the two class means are those with a high eigenvalue $\lambda_d$ and an eigenvector angularly aligned to the vector $\mathbf{e}_k - \mathbf{e}_m$.

We express the weighted pair-wise distance between all $C$ classes in $\mathcal{S}$ by:
\begin{eqnarray}
\mathcal{D} &=& \sum_{k=1}^{C} \sum_{m=1}^{C} p_k p_m \: d(c_k,c_m) \nonumber \\
&=& \sum_{k=1}^{C} \sum_{m=1}^{C} \sum_{d=1}^{N}  \lambda_d p_k p_m \left( \mathbf{u}_d^T \mathbf{e}_k - \mathbf{u}_d^T \mathbf{e}_m \right)^2 \nonumber \\
&=&  \sum_{d=1}^N \lambda_d \: D_d \label{Eq:WeightedDistanceClassPairs}
\end{eqnarray}
where each class contributes proportionally to its cardinality $p_c = N_c / N, \:c=1,\dots,C$ and
\begin{equation}
D_d = \frac{1}{N^2} \sum_{k=1}^{C} \sum_{m=1}^{C} (N_k + N_m) \cos^2(\mathbf{u}_d,\mathbf{e}_k-\mathbf{e}_m)
\end{equation}
expresses the weighted alignment of the eigenvector $\mathbf{u}_d$ to all possible combinations of class indicator vectors difference.

To define the subspace $\mathbb{R}^M$ of the kernel space $\mathcal{F}$ that maximally preserves the pair-wise distances between the class means in the kernel space, we keep the $M$ eigen-pairs minimizing:
\begin{equation}
\Delta D = (\mathcal{D} - \mathcal{D}_{1:M} )^2 \label{Eq:Criterion}
\end{equation}
where $\mathcal{D}_{1:M}$ is defined as the weighted pair-wise distance between all $C$ classes in $\mathcal{S}$ when using the $M$ selected eigen-pairs, i.e. using (\ref{Eq:WeightedDistanceClassPairs}):
\begin{equation}
D_{1:M} = \sum_{d=1}^N \alpha_d \lambda_d \: D_d \label{Eq:WeightedDistanceClassPairsSelected}
\end{equation}
where $\alpha_d = 1$ if dimension $d$ is selected and $\alpha_d = 0$ otherwise.

Thus, in contrary to (uncentered) kPCA and kECA selecting the eigen-pairs corresponding to the maximal eigenvalues or entropy values, for the selection of an eigen-pair in CMVCA both the eigenvalue $\lambda_d$ needs to be high and the corresponding eigenvector needs to be angularly aligned to the difference of a pair of class indicator vectors.

\subsection{CMVCA preserves the class means to total mean distances}\label{SS:CMVAmeanVectors}
In the above we defined CMVCA as the method preserving the pair-wise distances between class means in $\mathcal{F}$. Considering the weighted distance value of dimension $d$ from (\ref{Eq:WeightedDistanceClassPairs}), and by exploiting that $\sum_{m=1}^C p_m = 1$ and $\mathbf{e} = \sum_{c=1}^C p_c \mathbf{e}_c$, we have:
\begin{eqnarray}
D_d &=& \sum_{k=1}^{C} \sum_{m=1}^{C} p_k p_m \left( (\mathbf{u}_d^T \mathbf{e}_k)^2 -2 \mathbf{u}_d^T \mathbf{e}_k \mathbf{u}_d^T \mathbf{e}_m + (\mathbf{u}_d^T \mathbf{e}_m)^2 \right) \nonumber \\
&=& 2 \sum_{k=1}^C p_k (\mathbf{u}_d^T \mathbf{e}_k)^2  - 2 \sum_{k=1}^{C} \sum_{m=1}^{C} p_k p_m \mathbf{u}_d^T \mathbf{e}_k \mathbf{u}_d^T \mathbf{e}_m \nonumber \\
&=& 2 \Bigg( \sum_{k=1}^C p_k (\mathbf{u}_d^T \mathbf{e}_k)^2 - 2 \sum_{k=1}^{C} \sum_{m=1}^{C} p_k p_m \mathbf{u}_d^T \mathbf{e}_k \mathbf{u}_d^T \mathbf{e}_m   \nonumber \\
&&+ \sum_{k=1}^C p_k \mathbf{u}_d^T \mathbf{e}_k \left( \sum_{m=1}^C p_m \mathbf{u}_d^T \mathbf{e}_m \right) \Bigg) \nonumber \\
&=& 2 \Bigg( \sum_{k=1}^C p_k \left( (\mathbf{u}_d^T \mathbf{e}_k)^2 -2 \mathbf{u}_d^T \mathbf{e}_k \mathbf{u}_d^T \mathbf{e} + (\mathbf{u}_d^T \mathbf{e})^2 \right) \Bigg) \nonumber \\
&=& 2 \sum_{k=1}^C p_k (\mathbf{u}_d^T \mathbf{e}_k - \mathbf{u}_d^T \mathbf{e})^2,  \label{Eq:WeightedDistanceClasses}
\end{eqnarray}
where $\mathbf{e} \in \mathbb{R}^N$ is a vector having all its elements equal to $1/N$. Combining (\ref{Eq:WeightedDistanceClasses}) with (\ref{Eq:WeightedDistanceClassPairs}) we obtain:
\begin{eqnarray}
\mathcal{D} &=& \sum_{d=1}^N \lambda_d \: D_d = 2 \sum_{d=1}^{N} \sum_{k=1}^{C} \lambda_d \: p_k \: (\mathbf{u}_d^T \mathbf{e}_k - \mathbf{u}_d^T \mathbf{e})^2 \label{Eq:DistanceTotalMean1}\\ \nonumber 
&=& 2 \sum_{d,k=1}^{N,C} p_k \: (\sqrt{\lambda_d}\mathbf{u}_d^T \mathbf{e}_k - \sqrt{\lambda_d}\mathbf{u}_d^T \mathbf{e})^2  \nonumber \\
&=& 2 \sum_{k=1}^C p_k \|\mathbf{m}_k - \mathbf{m}\|_2^2  \label{Eq:DistanceTotalMean}
\end{eqnarray}
where $\mathbf{m}$ is the total mean vector in $\mathcal{F}$. Thus, the eigen-pairs selected by minimizing the criterion in (\ref{Eq:Criterion}) are those preserving the distances between the class means to the total mean in $\mathcal{F}$.

\subsection{CMVCA as the Euclidean divergence between the class probability density functions}\label{SS:CMVAkernelEntropy}
Let us assume that the data forming $\mathcal{S}_k$ and $\mathcal{S}_m$ are drawn from the probability density functions $p_k(\mathbf{x})$ and $p_m(\mathbf{x})$, respectively. The Euclidean divergence between these two probability density functions is given by:
\begin{equation}
D(p_k,p_m) = \int p^2_k(\mathbf{x}) d\:\mathbf{x} - 2 \int p_k(\mathbf{x})p_m(\mathbf{x}) d\:\mathbf{x} + \int p^2_k(\mathbf{x}) d\:\mathbf{x}.
\end{equation}
Given the observations of these two probability density functions in $\mathcal{S}_k$ and $\mathcal{S}_m$, $D(p_k,p_m)$ can be estimated using the Parzen window method \cite{principe2010information,williams2000effect}. Let $\kappa_{\sigma}(\mathbf{x}_i,\cdot)$ be the Gaussian kernel centered at $\mathbf{x}_i$ with width $\sigma$. Then, we have:
\begin{eqnarray}
\hat{D}(p_k,p_m) &=& \frac{1}{N_k^2} \sum_{\mathbf{x}_i \in \mathcal{S}_k} \sum_{\mathbf{x}_j \in \mathcal{S}_k} \kappa_{\sigma}(\mathbf{x}_i,\mathbf{x}_j) \nonumber \\
&&- \frac{2}{N_k N_m} \sum_{\mathbf{x}_i \in \mathcal{S}_k} \sum_{\mathbf{x}_j \in \mathcal{S}_m} \kappa_{\sigma}(\mathbf{x}_i,\mathbf{x}_j) \nonumber \\
&&+ \frac{1}{N_m^2} \sum_{\mathbf{x}_i \in \mathcal{S}_m} \sum_{\mathbf{x}_j \in \mathcal{S}_m} \kappa_{\sigma}(\mathbf{x}_i,\mathbf{x}_j) \nonumber \\
&=& \mathbf{e}_k^T \mathbf{K} \mathbf{e}_k - 2 \mathbf{e}_k^T \mathbf{K} \mathbf{e}_m  + \mathbf{e}_m^T \mathbf{K} \mathbf{e}_m
\end{eqnarray}
or expressing it using the eigen-decomposition of $\mathbf{K}$:
\begin{equation}
\hat{D}(p_k,p_m) = \sum_{d=1}^N \lambda_d \left( \mathbf{u}_d^T \mathbf{e}_k - \mathbf{u}_d^T \mathbf{e}_m \right)^2.
\end{equation}
Note here that the estimated Euclidean divergence between $p_k(\mathbf{x})$ and $p_m(\mathbf{x})$ gets the same form as the distance of the class mean vectors of classes $c_k$ and $c_m$ in (\ref{Eq:MeanVectorsDistancePairs}). Thus, $\mathcal{D}$ in (\ref{Eq:WeightedDistanceClassPairs}) can be expressed as:
\begin{equation}
\mathcal{D} = \sum_{k=1}^{C} \sum_{m=1}^{C} p_k p_m \: \hat{D}(p_k,p_m). \label{Eq:WeightedEntropyPairs}
\end{equation}
From the above, it can be seen that the dimensions minimizing the criterion in (\ref{Eq:Criterion}), are those optimally preserving the weighted pair-wise Euclidean divergence between the probability density functions of the classes in the input space. Interestingly, exploiting the PSD property of the kernel matrix, the analysis in \cite{jenssen2011kernel} based on the expected value of kernel convolution operator shows that the Parzen window method can be replaced by any PSD kernel function. Our results are complementary to those presented in \cite{ran2018connections} studying the connection of R\'enyi entropy PCA and kernel learning.

\subsection{CMVCA in terms of uncentered PCA projections}\label{SS:CMVAuncenteredPCA}
Let us denote by $\mathbf{v}_d$ the $d$-th eigenvector of the scatter matrix $\mathbf{S}_T^{(\phi)} = \tilde{\mathbf{\Phi}} \tilde{\mathbf{\Phi}}^T$ of the data. $\mathbf{v}_d$ is in essence a projection vector defined by applying uncentered kernel PCA on the input vectors $\mathbf{x}_i, \:i=1,\dots,N$. By using the connection between the eigenvectors of $\mathbf{S}_T^{(\phi)}$ and $\mathbf{K}$ \cite{wermuth1993eigenanalysis}, we substitute $\mathbf{u}_d^T = \frac{1}{\sqrt{\lambda_d}} \mathbf{v}_d^T \tilde{\mathbf{\Phi}}$ in \ref{Eq:DistanceTotalMean1} and $\mathcal{D}$ becomes: 
\begin{eqnarray}
\mathcal{D} &=& 2 \sum_{k=1}^{C} \sum_{d=1}^{N} p_k  (\mathbf{v}_d^T \tilde{\mathbf{\Phi}} \mathbf{e}_k - \mathbf{v}_d^T \tilde{\mathbf{\Phi}} \mathbf{e})^2 \nonumber \\
&=& 2 \sum_{d=1}^N  \left( \sum_{k=1}^C p_k \Big(\mathbf{v}_d^T (\mathbf{m}_k - \mathbf{m}) \Big)^2 \right) = 2 \sum_{d=1}^N  \hat{D}_d.  \label{Eq:Dhat_OvR}
\end{eqnarray}
Using $\|\mathbf{v}_d\|_2^2 = 1$, we get:
\begin{equation}
\mathcal{D} = 2 \sum_{k=1}^C p_k \|\mathbf{m}_k - \mathbf{m}\|_2^2 \left( \sum_{d=1}^N \cos^2(\mathbf{v}_d, \mathbf{m}_k - \mathbf{m}) \right). \label{Eq:D_OvR}
\end{equation}
Since $\sum_{d=1}^N \cos^2(\mathbf{v}_d, \mathbf{m}_k - \mathbf{m}) = 1$, the contribution of uncentered kernel PCA axis $\mathbf{v}_d$ to $\|\mathbf{m}_k - \mathbf{m}\|_2^2$ is determined by the cosine of the angle between $\mathbf{v}_d$ and $\mathbf{m}_k - \mathbf{m}$ in the sense that the axes which are most angularly aligned with $\mathbf{m}_k - \mathbf{m}$ contribute the most. This result adds to the insight provided in \cite{cadima2009relationships,jenseen2013mean} and defines CMVCA in terms of the projections obtained by applying uncentered kernel PCA on the input data.

\subsection{Connection between CMVCA and KDA}\label{SS:CMVAuncenteredLDA}
To analyze the connection between CMVCA and KDA, we define the within-class scatter matrix:
\begin{equation}
\mathbf{S}_w^{(\phi)} = \sum_{k=1}^C \sum_{\mathbf{\phi}_i \in \mathcal{S}_k} (\mathbf{\phi}_i - \mathbf{m}_k)(\mathbf{\phi}_i - \mathbf{m}_k)^T \label{Eq:Dw_kda}
\end{equation}
and the between-class scatter matrix:
\begin{equation}
\mathbf{S}_b^{(\phi)} =\sum_{k=1}^C N_k (\mathbf{m}_k - \mathbf{m})(\mathbf{m}_k - \mathbf{m})^T. \label{Eq:Db_kda}
\end{equation}
The total distance is then given by $\mathbf{S}_T^{(\phi)} = \mathbf{S}_w^{(\phi)} + \mathbf{S}_b^{(\phi)}$, i.e:
\begin{equation}
\mathbf{S}_T^{(\phi)} = \sum_{i=1}^N (\mathbf{\phi}_i - \mathbf{m})(\mathbf{\phi}_i - \mathbf{m})^T. \label{Eq:ST_kda}
\end{equation}

Using the above scatter matrices, KDA and its variants \cite{iosifidis2013optimal,iosifidis2014krda} the eigenvectors $\mathbf{v}_d$ maximizing the Rayleigh quotient:
\begin{equation}
\mathbf{V}^{*} = \underset{\mathbf{V}^T\mathbf{V}=\mathbf{I}}{\arg \max} \: \frac{ Tr \left( \mathbf{V}^T \mathbf{S}_b^{(\phi)} \mathbf{V} \right) }{ Tr \left( \mathbf{V}^T \mathbf{S}_T^{(\phi)} \mathbf{V} \right) },
\end{equation}
leading to at most $C-1$ axes which are the eigenvectors corresponding to the positive eigenvalues of the generalized eigen-problem $\mathbf{S}_b^{(\phi)} \mathbf{v} = \lambda \mathbf{S}_T^{(\phi)} \mathbf{v}$.

Here we are interested in the discrimination power in terms of the KDA criterion of the axes $\mathbf{v}_d, \:d=1,\dots,N$ defined from the spectral decomposition of $\mathbf{K}$. Expressing the above projections based on the eigenvectors of $\mathbf{S}_T^{(\phi)}$ and assuming the data to be centered, i.e., $\mathbf{m} = \mathbf{0}$, we have $\mathbf{v}_d^T \tilde{\mathbf{\Phi}}\tilde{\mathbf{\Phi}}^T \mathbf{v}_d = \lambda_d$. By using $p_k = N_k / N$ and $\mathbf{m}_k = \tilde{\mathbf{\Phi}} \mathbf{e}_k$ the Rayleigh quotient for axis $d$ is equal to:
\begin{eqnarray}
\frac{ \mathbf{v}_d^T \mathbf{S}_b^{(\phi)} \mathbf{v}_d }{ \mathbf{v}_d^T \mathbf{S}_T^{(\phi)} \mathbf{v}_d } &=& \frac{ \mathbf{v}_d^T \left( \sum_{k=1}^C N_k \mathbf{m}_k\mathbf{m}_k^T \right) \mathbf{v}_d }{ \mathbf{v}_d^T \tilde{\mathbf{\Phi}}\tilde{\mathbf{\Phi}}^T \mathbf{v}_d } \nonumber \\
&=& \sum_{k=1}^C N_k \left( \frac{1}{\sqrt{\lambda_d}}\mathbf{v}_d^T \tilde{\mathbf{\Phi}} \mathbf{e}_k \right)^2 = \sum_{k=1}^C N_k \left( \mathbf{u}_d^T \mathbf{e}_k \right)^2 \nonumber \\
&=& \sum_{k=1}^C N_k \Big( \|\mathbf{u}_d\|_2 \|\mathbf{e}_k\|_2 \: cos(\mathbf{u}_d,\mathbf{e}_k) \Big)^2 \nonumber \\
&=& \sum_{k=1}^C \frac{1}{N_k} \cos^2(\mathbf{u}_d,\mathbf{e}_k). \label{Eq:KDAdimension}
\end{eqnarray}

The criterion of CMVCA from (\ref{Eq:WeightedDistanceClassPairs}) and (\ref{Eq:WeightedDistanceClasses}) for axis $d$ becomes:
\begin{equation}
\mathcal{D}_d = \lambda_d D_d = 2 \sum_{k=1}^C p_k \lambda_d (\mathbf{u}_d^T \mathbf{e}_k)^2 = \frac{2}{N} \sum_{k=1}^C \frac{\lambda_d}{N_k} \cos^2(\mathbf{u}_d,\mathbf{e}_k).  \label{Eq:CVMAdimensionKDA}
\end{equation}
Thus, while in CMVCA an eigen-pair contributes to the criterion based on both the size of $\lambda_d$ and the angular alignment between $\mathbf{u}_d$ and the class indicator vectors $\mathbf{e}_k$, the criterion of KDA selects dimensions based on only the angular alignment between $\mathbf{u}_d$ and the class indicator vectors $\mathbf{e}_k$. Note that (\ref{Eq:KDAdimension}) also gives new insights on why the KDA criterion restricts the dimensionality of the produced subspace by the number of classes. That is, since by definition $\mathbf{u}_d$ form an orthogonal basis, the number of eigen-vectors that can be angularly aligned to the class indicator vectors is restricted by the number of classes $C$, which is equal to the rank of the between-class scatter matrix for uncentered data. We will exploit this observation to define a discriminative version of CMVCA in the following.

\section{Class Mean Vector Discriminant Analysis}\label{SS:DiscriminantCVMA}
An interesting extension of CMVCA is the CMVDA, which motivated by the connection of CMVCA with KDA obtained by following the analysis in Subsection \ref{SS:CMVAuncenteredLDA}. By comparing (\ref{Eq:KDAdimension}) and (\ref{Eq:CVMAdimensionKDA}) we see that in the case where $\lambda_d = 1, \:d=1,\dots,N$, the scores calculated for the kernel subspace dimensions by CMVCA and KDA are the same. This situation arises when the data $\tilde{\mathbf{\Phi}}$ is whitened, i.e. when $\mathbf{S}_T^{(\phi)} = \tilde{\mathbf{\Phi}} \tilde{\mathbf{\Phi}}^T = \mathbf{I}$, where $\mathbf{I} \in \mathbb{R}^{N \times N}$ is the identity matrix. Interestingly, the information needed for whitening $\tilde{\mathbf{\Phi}}$ can be directly obtained by the eigenanalysis of $\mathbf{K} = \tilde{\mathbf{\Phi}}^T \tilde{\mathbf{\Phi}}$, since there is a direct connection between the eigenvalues and eigenvectors of $\mathbf{S}_T^{(\phi)}$ and $\mathbf{K}$ \cite{wermuth1993eigenanalysis}.

Given a kernel matrix $\tilde{\mathbf{K}} = \tilde{\mathbf{\Phi}}^T \tilde{\mathbf{\Phi}}$, where $\tilde{\mathbf{\Phi}}$ is whitened, application of CMVCA requires eigen-decomposition of $\tilde{\mathbf{K}}$ for calculating the eigen-vectors $\mathbf{u}_d, \:d=1,\dots,N$ and the corresponding eigenvalues $\lambda_d$ to be used for weighting the dimensions of the kernel subspace based on (\ref{Eq:WeightedDistanceClassPairs}). $\tilde{\mathbf{K}}$ has all its eigenvalues equal to $\lambda_d = 1,\:d=1,\dots,N$ and, thus, its eigenvectors form the axes of an arbitrary basis, i.e.:
\begin{equation}
\{\mathbf{u}_d\}_{d=1}^N, \:\:\: \left( \mathbf{u}_i^T \mathbf{u}_i = 1, \:\:\: \mathbf{u}_i^T \mathbf{u}_j = 0, \:i \neq j \right). \label{Eq:RandomBasis}
\end{equation}
Such basis can be efficiently calculated by applying an orthogonalization process (e.g. Cholesky decomposition) starting from a vector belonging to the span of $\tilde{\mathbf{\Phi}}$. The vector $\mathbf{e}$ is such a vector and, thus, can be used for generating the basis.

Moreover the vectors $\sqrt{N_c}\:\mathbf{e}_c, \:c=1,\dots,C$ belong to the span of $\tilde{\mathbf{\Phi}}$ and also satisfy the two properties, i.e., $N_k \mathbf{e}_k^T \mathbf{e}_k = 1$ and $\sqrt{N_k N_m}\:\mathbf{e}_k^T \mathbf{e}_m = 0, \:k \neq m$. Thus, they can be selected to form the first $C$ eigenvectors of $\tilde{\mathbf{K}}$. Note that from (\ref{Eq:CVMAdimensionKDA}) it can be seen that these vectors contribute the most to the Rayleigh quotient criterion. To form the rest $N-C$ bases, we can apply an orthogonalization process on the subspaces determined by each class indicator vector $\mathbf{e}_c$, each generating a basis in $\mathbb{R}^{N_c}$ appended by zeros for the remaining dimensions, leading to $\sum_{c=1}^C N_c = N$ eigenvectors in total.

\section{Approximate kernel subspace learning}\label{SS:CMVAapproxKernels}
In cases where the cardinality of $\mathcal{S}$ is prohibitive for applying kernel methods, approximate kernel matrix spectral analysis can be used. Probably the most widely used approach is based on the Nystr\"{o}m method, which first chooses a set of $n << N$ reference vectors to calculate the kernel matrix between the reference vectors $\mathbf{K}_{nn} \in \mathbb{R}^{n \times n}$ and the matrix $\mathbf{K}_{Nn} \in \mathbb{R}^{N \times n}$ containing the kernel function values between the training and reference vectors. In order to determine the reference vectors, two approaches have been proposed. The first is based on selecting $n$ columns of $\mathbf{K}$ using random or probabilistic sampling \cite{williams2001using,kumar2009sampling}, while the second one determines the reference vectors by applying clustering on the training vectors \cite{zhang2010clustered,iosifidis2016anpt}.

The Nystr\"{o}m-based approximation of $\mathbf{K}$ is given by
\begin{equation}
    \mathbf{K} \simeq \mathbf{K}_{Nn} \mathbf{K}_{nn}^{-1} \mathbf{K}_{Nn}^T = \tilde{\mathbf{\Phi}}_n^T \tilde{\mathbf{\Phi}}_n,    
\end{equation}
where $\tilde{\mathbf{\Phi}}_n = \mathbf{K}_{nn}^{-\frac{1}{2}} \mathbf{K}_{Nn}^T \in \mathbb{R}^{n \times N}$. When the ranks of $\mathbf{K}$ and $\mathbf{K}_{nn}$ match, this gives an exact calculation of $\mathbf{K}$ and $\tilde{\mathbf{\Phi}}_n$ is the same as $\tilde{\mathbf{\Phi}}$ defined in Section \ref{S:Preliminaries}. Eigen-decomposition of $\mathbf{K}_{nn}$ leads to $\mathbf{K}_{nn}^{-\frac{1}{2}} = \mathbf{U}_n \mathbf{\Lambda}_n^{-\frac{1}{2}} \mathbf{U}_n^T$. When $\mathbf{K}_{nn}$ is a $n$-rank matrix, the matrices $\tilde{\mathbf{\Phi}}_n^T \tilde{\mathbf{\Phi}}_n$ and $\tilde{\mathbf{\Phi}}_n\tilde{\mathbf{\Phi}}_n^T$ have the same $n$ leading eigenvalues \cite{wermuth1993eigenanalysis}. The matrix $\tilde{\mathbf{\Phi}}_n\tilde{\mathbf{\Phi}}_n^T$ is an $n \times n$ matrix and, thus, applying eigen-analysis to it can highly reduce the computational complexity required for the calculation of eigenvalues $\mathbf{\Lambda}_{(n)}$ of the approximate kernel matrix. Finally, the data representations in the approximate kernel subspace \cite{iosifidis2016anpt} are calculated by:
\begin{equation}
    \tilde{\mathbf{\Phi}}_n = \mathbf{\Lambda}_{(n)}^{-\frac{1}{2}} \mathbf{\Lambda}_n^{-1}\mathbf{U}_n^T \mathbf{K}_{Nn}^T \mathbf{K}.
\end{equation}
Another approach proposed for making the use of kernel methods in big data feasible is based on explicit nonlinear mappings. A nonlinear mapping $\mathbf{x}_i \in \mathbb{R}^D \rightarrow \mathbf{z}_i \in \mathbb{R}^n$ is defined such that $\mathbf{z}_i^T \mathbf{z}_j \simeq \kappa(\mathbf{x}_i,\mathbf{x}_j)$, or by using the entire dataset $\mathbf{Z}_n^T \mathbf{Z}_n \simeq \mathbf{K}$ \cite{rahimi2007random,ring2016approximation}. 

After the calculation of the data representations in $\mathbb{R}^n$ either by using the Nystr\"{o}m method or the randomized nonlinear mappings, we can apply the proposed CMVCA by applying singular value decomposition. That is, the right singular vectors corresponding to the non-zero singular values of either of the matrices $\tilde{\mathbf{\Phi}}_n$ or $\mathbf{Z}_n$ define the axes to be considered for minimizing the CMVCA criterion in (\ref{Eq:WeightedDistanceClassPairs}). In order to apply CMVDA on the approximate and randomized kernel cases the sample representations in the $\mathbb{R}^n$ are whitened and we follow the process described in Section \ref{SS:DiscriminantCVMA} to determine the eigen-pairs used for CMVDA-based projection.
\begin{figure*}
\centering
    \includegraphics[width=0.3\linewidth]{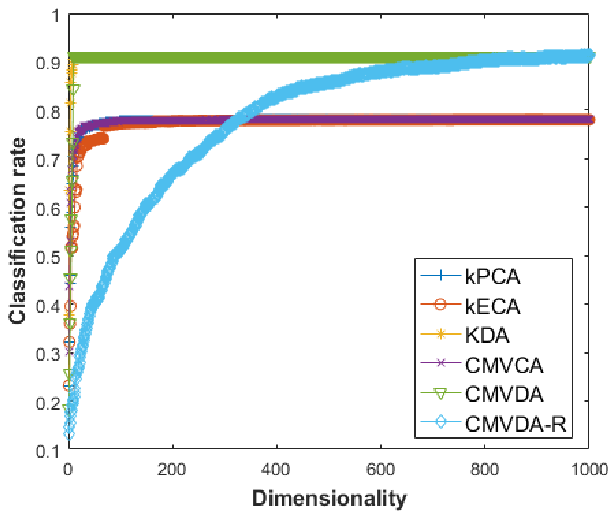}
    \includegraphics[width=0.3\linewidth]{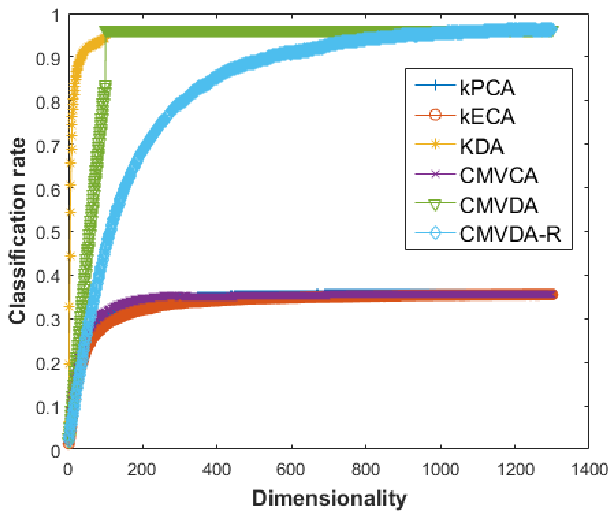}  
    \includegraphics[width=0.3\linewidth]{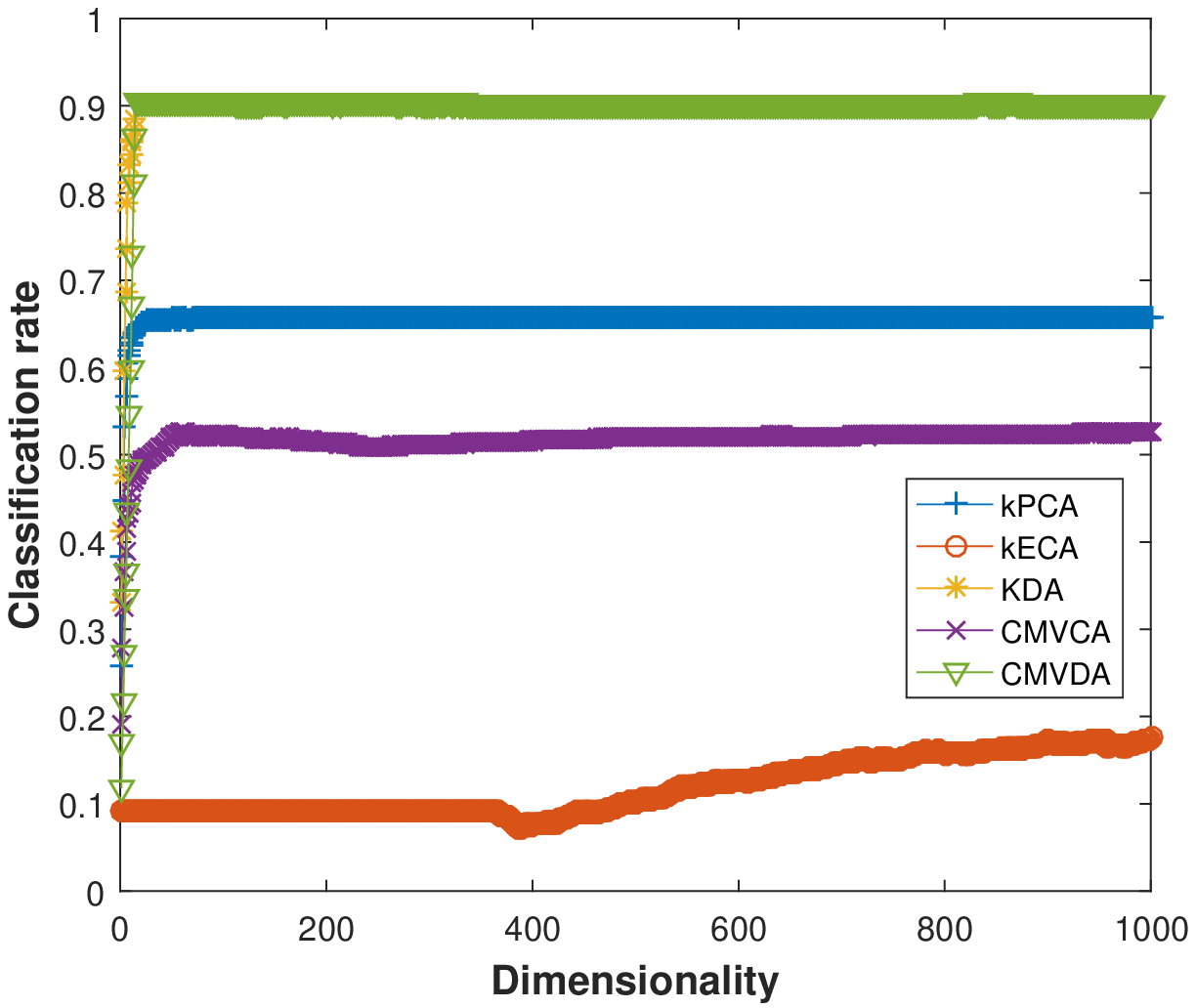} \\ \vspace{0.05cm}
    \includegraphics[width=0.3\linewidth]{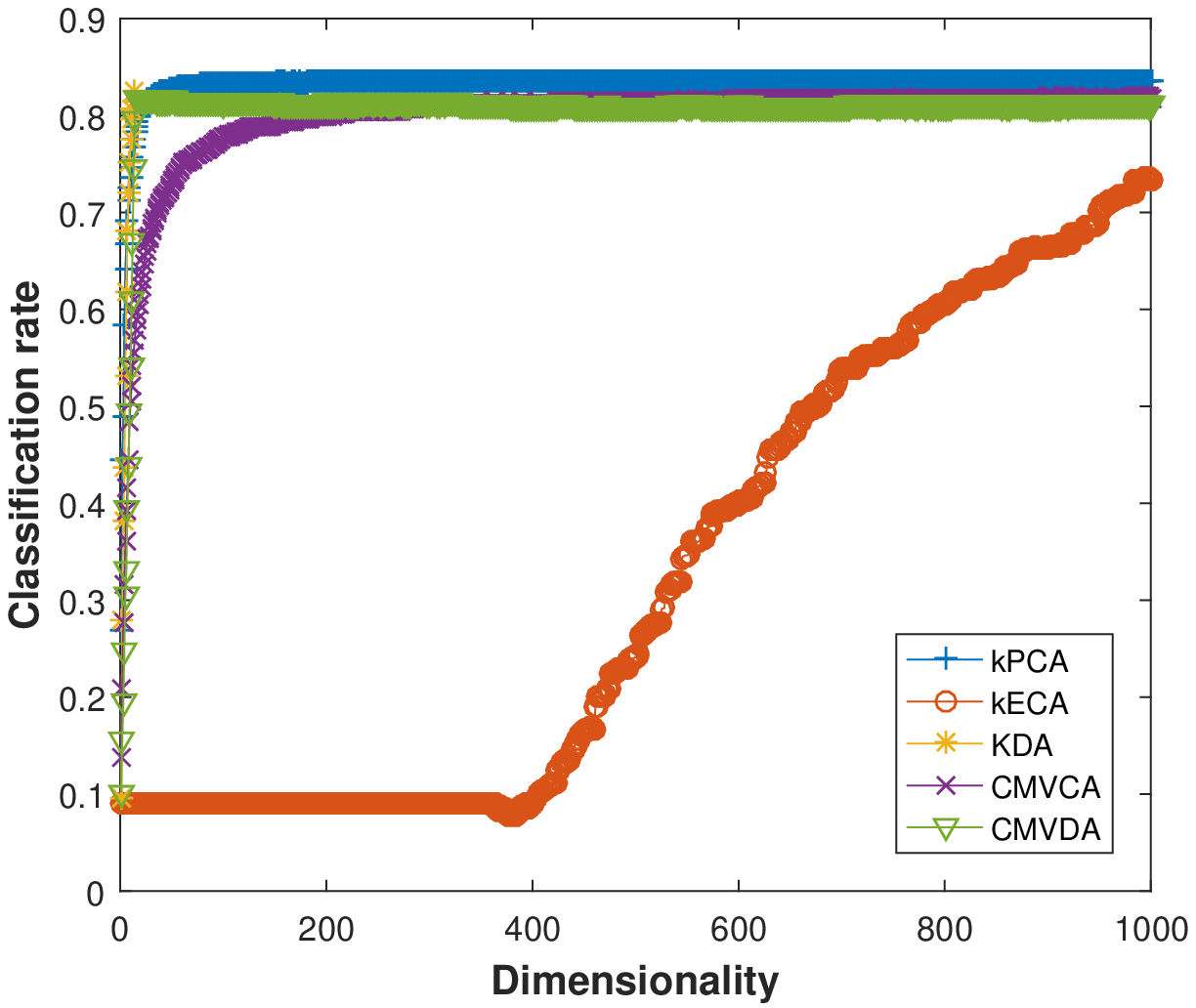}  
    \includegraphics[width=0.3\linewidth]{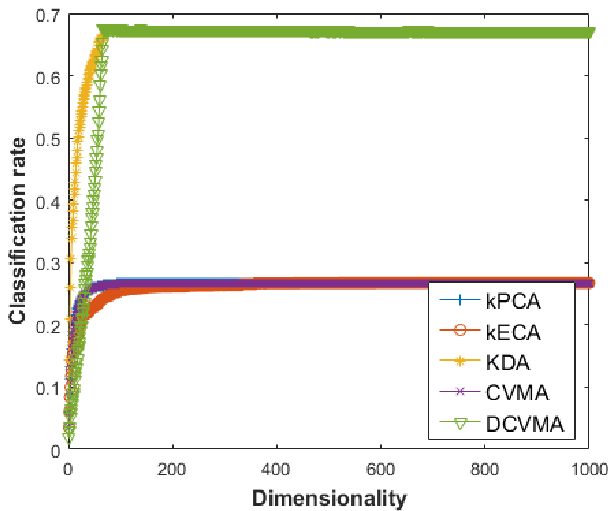}
    \includegraphics[width=0.3\linewidth]{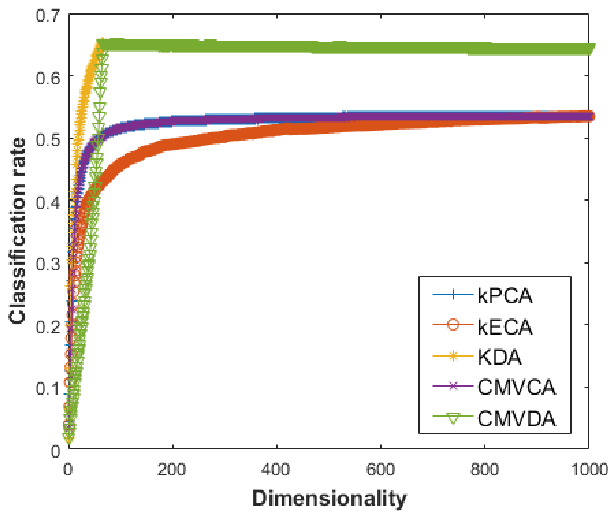}
    \caption{Classification rate vs. subspace dimensionality. From top-left to bottom-right: MNIST-100, AR, $15$ scene using Nystr\"{o}m approximation, $15$ scene using random features, MIT indoor using Nystr\"{o}m approximation and MIT indoor using random features.}
    \label{fig:videoframes}
\end{figure*}

\begin{figure}
\centering 
\includegraphics[width=0.49\linewidth]{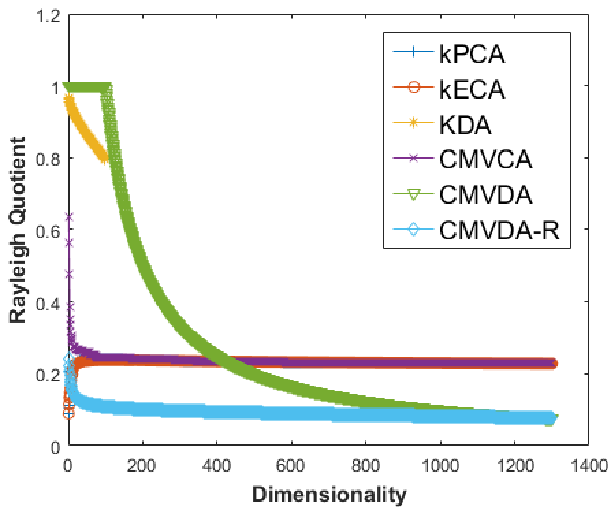}
\includegraphics[width=0.49\linewidth]{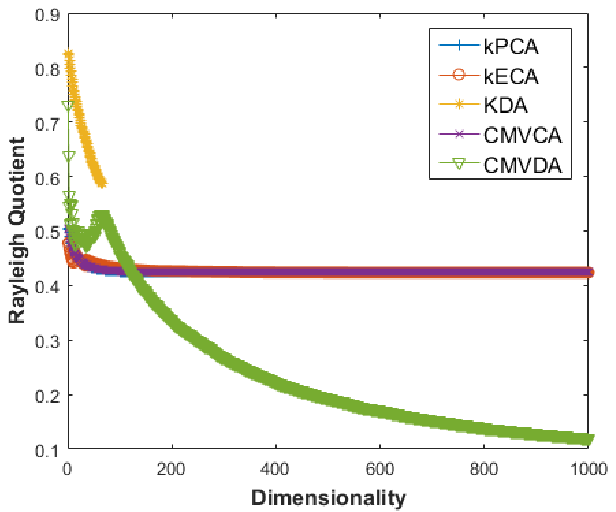}
\caption{\it Rayleigh quotient vs. subspace dimensionality on the (left) AR and (right) MIT indoor datasets.}\label{fig:Example}
\end{figure}
\begin{table}
\begin{center}
\caption{Datasets used in experiments.}\label{tbl:datasets}
\resizebox{0.6\linewidth}{!}{
\begin{tabular}{|l|c|c|c|} \hline
Dataset    &  $D$   &  $C$  & $N$      \\ \hline
MNIST-100  &  $784$ & $10$  & $1000$   \\ \hline
AR         & $1200$ & $100$ & $2600$   \\ \hline
$15$ scene &  $512$ & $15$  & $4485$   \\ \hline
MIT-indoor &  $512$ & $67$  & $15620$  \\ \hline
\end{tabular}}
\end{center}
\end{table}

\section{Experiments}\label{S:Experiments}
In our experiments we used four datasets, namely the MNIST \cite{lecun1998mnist}, AR \cite{martinez2001pca}, $15$ scene \cite{lazebnik2006beyond} and MIT indoor \cite{quattoni2009recognizing} datasets. For MNIST dataset, we used the first $100$ training samples per class to form the training set and we report performance on the entire test set. For the rest datasets, we perform ten experiments by randomly keeping half of the samples per class for training and the remaining for evaluation, and we report the average performance over the ten experiments. We used the vectorized pixel intensity values for representing images in MNIST and AR datasets. For the $15$ scene and MIT indoor datasets we used deep features generated by average pooling over spatial dimension of the last convolution layer of VGG network \cite{simonyan2014very} trained on ILSVRC2012 database, and we follow the approximate kernel and randomized kernel approaches using $n = 1000$. Details of these datasets are shown in Table \ref{tbl:datasets}.

In all experiments we used the Gaussian kernel function:
\begin{equation}
    [\mathbf{K}]_{ij} = exp\left( -\frac{ \|\mathbf{x}_i - \mathbf{x}_j \|_2^2 }{2 \sigma^2} \right)
\end{equation}
and set the value of $\sigma$ equal to the mean pair-wise distance between all training samples. In order to illustrate the effect of using different subspace dimensionality, we used the nearest class centroid classifier for all possible subspaces produced by each of the methods. This means that we applied each of the methods using the training data and obtain all projection vectors in the order determined by the corresponding criterion. Then we measure the performance of each method on all subspaces it determines as follows: For each projection dimensionality, we map the data to the kernel subspaces determined by all methods. We calculate the class means in the corresponding subspace using the training data and we perform nearest class centroid classification on the test data. This classifier was selected since it is the simplest linear classifier. This allows us to compare the (nonlinear) subspace learning methods in a more fair manner, compared to using other classifier choices. All experiments were conducted on a PC with i5 CPU at 2.3 GHz and 12GB RAM, using MATLAB 2016a.

Figure \ref{fig:videoframes} illustrates the performance obtained by applying kPCA, kECA, KDA and the proposed CMVCA, CMVDA and the variant of CVMDA-R using the random basis of the kernel matrix produced by whitened kernel effective space (Eq. (\ref{Eq:RandomBasis})) as a function of the subspace dimensionality. The maximal performance obtained by each method is provided in Table \ref{tbl:performance} for completeness. Figure \ref{fig:Example} illustrates the Rayleigh quotient values as a function of the dimensionality of the subspace produced by all methods for the AR and MIT indoor datasets. As can be seen, the value of the Rayleigh quotient of the subspaces obtained by applying the unsupervised methods are, as expected, low. The subspaces obtained by KDA lead to a high value, which is gradually decreasing as more dimensions are added. This is expected, since the number of meaningful projections defined by the KDA criterion is restricted by the number of classes (due to the rank of $\mathbf{S}_b^{(\phi)}$ being equal to $C-1$). CMVDA leads to subspaces with a high Rayleigh quotient value which is gradually reduced, similarly to KDA. This is due to that, based on Eq. (\ref{Eq:CVMAdimensionKDA}), the subspace dimensions obtained by the CMVDA need to be angularly aligned with the class indicator vectors $\mathbf{e}_k, \:k=1,\dots,C$. Since the eigenvectors of $\mathbf{K}$ need to also be orthogonal, the number of projections which can be angularly aligned to the class indicator vectors is upper-bounded by the number of classes. Similar behaviors were observed for the rest of the datasets.

\section{Conclusion}\label{S:Conclusions}
In this paper, we proposed a component analysis method, called Class Mean Vector Component Analysis (CMVCA), for kernel-based dimensionality reduction preserving the distances between the class means in the kernel space. We provided an analysis of the proposed criterion which shows that it also determines the subspace dimensions preserving the distances between the class means to the total mean in the kernel space, as well as the Euclidean divergence of the class probability density functions in the input space. Moreover, we showed that the proposed criterion, while expressing different properties, has relations to the criteria used in (uncentered) kPCA, kECA and KDA. The latter connection leads to a discriminant analysis extension of the proposed method for multi-class problems, called Class Mean Vector Discriminant Analysis (CMVDA).
\begin{table}[!h]
\begin{center}
\caption{Classification rates ($\%$) over the various datasets.}\label{tbl:performance}
\resizebox{0.95\linewidth}{!}{
\begin{tabular}{|l|c|c|c||c|c|} \hline
Dataset        &  kPCA   &   kECA  &  CMVCA  &   KDA   &  CMVDA  \\ \hline
MNIST-100      & $78.07$ & $78.08$ & $78.08$ & $90.63$ & $91.28$ \\ \hline
AR             & $35.53$ & $35.55$ & $35.54$ & $94.42$ & $96.26$ \\ \hline
$15$ scene (N) & $65.63$ & $17.51$ & $52.60$ & $88.55$ & $90.58$ \\ \hline
MIT Indoor (N) & $26.75$ & $26.75$ & $26.75$ & $66.16$ & $67.46$ \\ \hline
$15$ scene (R) & $83.60$ & $73.60$ & $81.90$ & $82.57$ & $81.73$ \\ \hline
MIT Indoor (R) & $53.50$ & $53.50$ & $53.50$ & $65.37$ & $65.25$ \\ \hline
\end{tabular}}
\end{center}
\end{table}

The advantages of the proposed approach compared to kPCA, kECA and KDA, include a) a clear interpretation of the obtained projections, since (as detailed in Sections \ref{S:ProposedMethod} and \ref{SS:DiscriminantCVMA}) the projections obtained by the proposed approach need to be angularly aligned to the class indicator vectors defined based on the labels of the training data, b) the projections obtained by the proposed CMVDA method can be directly defined as scaled versions of the class indicator vectors leading to an efficient calculation of the kernel subspace, since their calculation trivial, and c) as was shown in Section \ref{SS:CMVAapproxKernels} the proposed methods can be easily combined with randomized and low-rank kernel matrix approximation approaches for performing kernel subspace learning on large scale data sets. Naturally, the proposed approach inherits all disadvantages of kernel-based methodologies related to their computational complexity (or their dependence on the low-rank decomposition and randomization approaches used to approximate the kernel matrix) for big data problems. Compared to kernel-based subspace learning methods exploiting local neighborhood information of the input data (e.g. \cite{yan2007graph}) the proposed method by defining its projections based on global variance information can be more susceptible to the existence of noisy data.

Interesting extensions of our analysis would include the analysis of the contribution of the kernel subspace dimensions in different problems successfully solved by kernel-based learning methods like one-class classification \cite{mygdalis2016graph,sohrab208subspace} and verification based on class-specific subspace learning models \cite{iosifidis2018probabilistic}.

\section*{Acknowledgement}
This work was supported by the European Union’s Horizon 2020 Research and Innovation Action Program under Grant 871449 (OpenDR). This publication reflects the authors’ views only. The European Commission is not responsible for any use that may be made of the information it contains.

\bibliographystyle{IEEEtran}
\bibliography{bibliography}

\end{document}